\documentclass{article}
\usepackage[preprint]{neurips_2024}
\usepackage[utf8]{inputenc}
\usepackage[T1]{fontenc}
\usepackage{hyperref}
\usepackage{url}
\usepackage{booktabs}
\usepackage{amsfonts}
\usepackage{nicefrac}
\usepackage{microtype}
\usepackage{xcolor}
\usepackage{graphicx}
\usepackage{enumitem}
\usepackage{comment}

\usepackage{multirow} % Required for \multirow
\usepackage{rotating} % Required for \rotatebox
\usepackage{graphicx} % Required for resizing tables

\title{Synthetic Data Can Mislead Evaluations: \\Membership Inference  as Machine Text Detection}

\author{%
Ali Naseh,  Niloofar Mireshghallah\\
University of Massachusettes Amherst, University of Washington \\
\texttt{anaseh@umass.edu}, \texttt{niloofar@cs.washington.edu}
}

\begin{document}

\maketitle

\begin{abstract}
Recent work shows membership inference attacks (MIAs) on large language models (LLMs) produce inconclusive results, partly due to difficulties in creating non-member datasets without temporal shifts. While researchers have turned to synthetic data as an alternative, we show this approach can be fundamentally misleading. Our experiments indicate that MIAs function as machine-generated text detectors, incorrectly identifying synthetic data as training samples regardless of the data source. This behavior persists across different model architectures and sizes, from open-source models to commercial ones such as GPT-3.5. Even synthetic text generated by different, potentially larger models is classified as training data by the target model. Our findings highlight a serious concern: using synthetic data in membership evaluations may lead to false conclusions about model memorization and data leakage. We caution that this issue could affect other evaluations using model signals such as loss where synthetic data substitutes for real-world samples.

\end{abstract}

\section{Introduction}
Membership inference attacks (MIAs) serve as a critical tool for examining privacy concerns in large language models (LLMs), particularly their capacity to memorize training data. The implications of such memorization extend beyond privacy to detecting copyright violations \citep{henderson2023copyright}, identifying test set contamination \citep{aerni2024evaluations}, and auditing the use of proprietary data in training. While MIAs aim to determine whether specific data points were used during model training—a capability crucial for regulatory compliance and public trust—recent work has revealed fundamental challenges in their application. Multiple studies demonstrate that current attacks perform barely better than random guessing on open-source models \citep{duan2024membership,das2024blind}, raising questions about their reliability. These concerns are compounded by methodological issues in existing evaluation protocols, which either rely on temporal shifts that introduce confounding distribution differences, or suffer from high n-gram overlap between members and non-members \citep{maini2024llm,zhang2024membership}. This has led to an ``evaluation crisis'' in membership inference, where current methods fail to provide meaningful signals about training data leakage \citep{aerni2024evaluations}.

Researchers have increasingly turned to synthetic data as a potential solution to these evaluation challenges, as it circumvents both temporal shifts and training set overlap concerns \citep{kazmi2024panoramia}. The use of synthetic data in MIA research extends beyond non-member construction—from training models on synthetic datasets to avoid copyright and privacy issues, to releasing synthetic versions of proprietary training data \citep{Khan2023Impact,Guepin2023Synthetic}. However, our analysis reveals a fundamental flaw in this approach: \textit{Certain MIAs consistently misclassify synthetic data as training members, suggesting these attacks function more as detectors of machine-generated text than as membership detectors}. This behavior raises critical questions about the validity of using synthetic data to evaluate model memorization and privacy leakage, particularly as synthetic data becomes increasingly prevalent in language model evaluation protocols.

This connection between MIAs and synthetic text detection is both methodological and intuitive. Both approaches analyze target model signals to distinguish between data subsets: MIAs separate training members from non-members, while machine-generated text detection identifies synthetic from human-written text. Consider their parallel techniques: perturbation-based approaches like the Neighborhood Attack~\citep{mattern2023membership} for membership inference mirror DetectGPT~\citep{mitchell2023detectgpt} for generated text detection, while likelihood-based methods like Min-k++\citep{zhang2024mink} parallel Fast-DetectGPT\citep{bao2024fast}. This equivalence is not coincidental—the signals that MIAs use, such as loss values and likelihood patterns, are precisely what language model sampling procedures optimize for, making synthetic data inherently similar to training data in the feature space these attacks examine~\citep{mireshghallah2023smaller}.

\begin{figure*}[t!]
\centering
  \label{fig:main_heatmap_checkpoint}
      \includegraphics[width=0.9\linewidth]{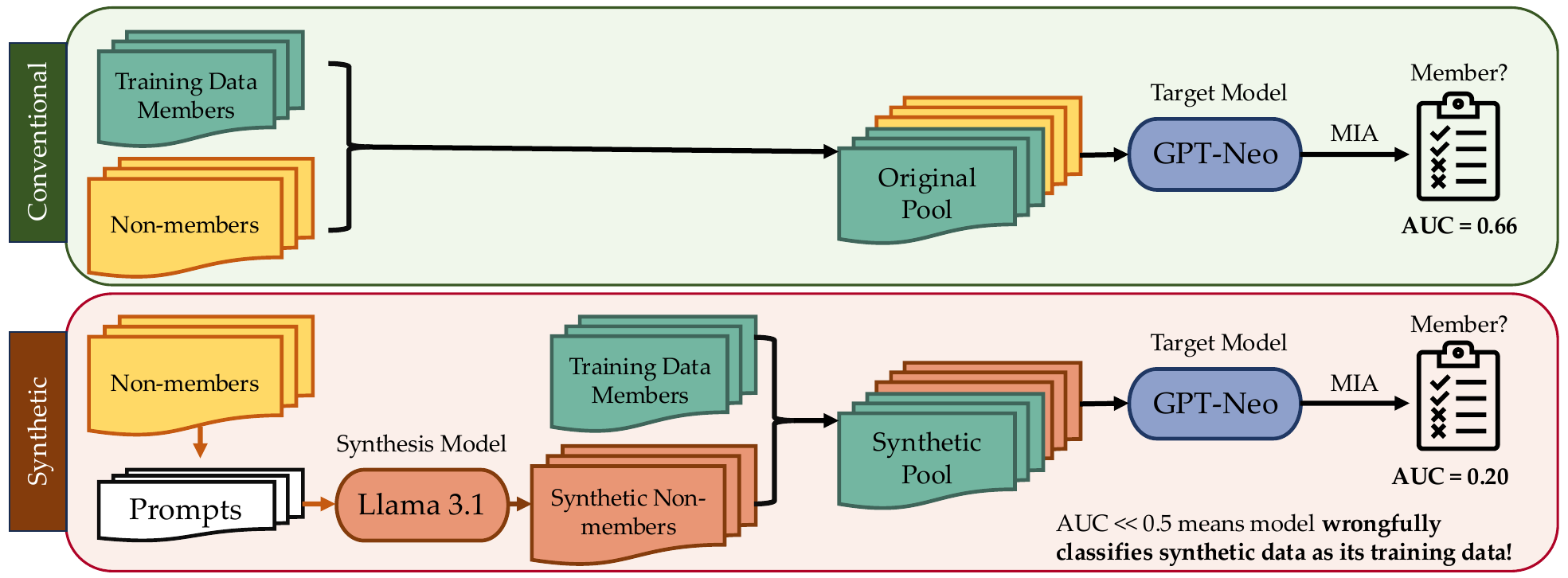}
\caption{Overview of our methodology: The conventional setup (top) evaluates MIAs using human-written members and non-members from MIMIR, while the synthetic setup (bottom) replaces non-members with machine-generated continuations, produced by prompting generator models with the first 30 tokens of each non-member. The AUC drops from 0.66 to 0.20 between setups, with AUC $\ll$ 0.5 indicating that MIAs consistently misclassify synthetic text as training data. Both setups use GPT-Neo 2.7B as the target model.}
     \vspace{-1ex }
     \label{fig:methodology}
\end{figure*}

While MIAs should assign higher membership scores to training members than non-members (synthetic or human), our experiments reveal the opposite. We demonstrate this phenomenon through two experimental setups (see Fig.~\ref{fig:methodology}): first, a conventional setup where we evaluate MIAs on human-written members and non-members from the MIMIR benchmark (top); second, a synthetic setup where we replace the non-members with machine-generated continuations of the same sequences. Using GPT-Neo 2.7B as our target model and LLaMA 3.1 as the generator, we find that the AUC drops dramatically from 0.66 to 0.20 when switching to synthetic non-members. Even with text generated by more capable models like GPT-3.5, the AUC remains below 0.5 (0.39)—indicating that MIAs consistently misidentify machine-generated text as training data, preferring synthetic generations over actual human-written training members. We validate these findings extensively across five different generator models, two data subsets, and five different MIAs, observing consistent patterns in all but one attack method (Zlib compression-based attack).

%This overlap raises a critical question: what happens when MIAs are tasked with distinguishing training members from synthetic text? This scenario arises directly when synthetic data is used as non-members in evaluations, mentioned earlier \citet{kazmi2024panoramia}. Intuitively, since synthetic data represents non-members, MIAs should assign higher scores to training members. Yet, contrary to expectation, our findings reveal that synthetic data systematically receives higher scores, exposing a fundamental limitation in the behavior of MIAs when applied in this context.
%
%To systematically study this phenomenon, we construct three distinct data pools: human-written training members from known pre-training datasets, human-written non-members matched for domain and complexity, and machine-generated text from various sources. We evaluate across multiple model scales and architectures, from open-source models (Pythia 6.9B, GPT-Neo 2.7B) to commercial APIs (ChatGPT), applying a comprehensive suite of membership inference attacks including loss-based, likelihood-based, and perturbation-based methods\ancomment{Shouldn't we mention these various models are the ones used to generate the machine-generated non-members? People might think we are talking about the target model. In our experiments, we are using one model as the target model}.

These findings have broader implications for language model evaluation beyond membership inference. Our work suggests a fundamental flaw in evaluations that rely on synthetic data. Many evaluation protocols leverage machine-generated text, from machine translation for cross-lingual assessment~\citep{wang2023evaluating} to language models judging other models' outputs~\citep{zhu2023judgelm} and other synthetic data training and evaluations~\citep{Guepin2023Synthetic}. Our results suggest such evaluations may be systematically biased—the signals they measure may be confounded by the synthetic nature of their data rather than the properties they aim to assess. This raises three critical questions: {\em (i)}~how does using language models to evaluate other models impact benchmark reliability, given their shared biases in processing synthetic text? {\em (ii)}~are synthetic data-based evaluations measuring intended properties or merely detecting machine generation artifacts? {\em (iii)}~why does synthetic text behavior transfer so consistently across different model scales and architectures? Recent work~\citep{mireshghallah2023smaller} suggests these patterns stem from fundamental similarities in how language models encode and process text.

\section{Membership Inference Attacks and Generated Text Detection}
\label{sec:background}

We argue that membership inference attacks (MIAs) and zero-shot machine-generated text detectors share surprisingly similar signals. Although their stated goals—identifying training set members versus identifying synthetic text—are distinct, both leverage the target model’s probability surface in comparable ways. Below, we explain each approach and illustrate how they converge on the same underlying likelihood cues.

\subsection{Membership Inference Attacks (MIAs)}
\label{sec:background_mia}
Membership inference attacks attempt to discern whether a particular sample was part of a model’s training set.
Broadly, MIA methods produce a score $f(x; M)$ indicating the likelihood that $x$ is a member of $M$’s training set, applying a threshold to yield a final prediction. Key variants include:
\begin{itemize}
    \item \textbf{Loss-Based Attack}~\citep{yeom2018privacy}: 
    Uses the model’s loss (or log-likelihood) on $x$ directly as a membership score, assuming that $x$ will incur a lower loss if it was part of training.
    \item \textbf{Reference-Based Attack}~\citep{carlini2021extracting}: 
    Introduces a second, comparable model (the ``reference'' model) to normalize the raw loss. The membership score compares how $M$ and the reference model each handle $x$, correcting for data or architectural differences.
    \item \textbf{Zlib Attack}~\citep{carlini2021extracting}:
    Normalizes the model’s loss by the zlib compression size of $x$. The idea is that compression length approximates text complexity or repetitiveness; inputs that compress poorly might elicit higher losses unless the model is trained specifically on similar data.
    \item \textbf{Min-K\%}~\citep{shi2023detecting}: 
    Sorts per-token likelihoods in ascending order and averages the lowest $K\%$. Non-members are hypothesized to have more low-likelihood tokens, thus yielding a higher average over the bottom $K\%$.
    \item \textbf{Min-K\%++}~\citep{zhang2024mink}: 
    Extends Min-K\% by normalizing token likelihoods based on their global mean and standard deviation. This approach further sharpens the contrast between member and non-member scores.
\end{itemize}

%Most methods assign a numerical score to each input (often based on loss or likelihood) and then use a threshold to predict membership. For instance, \textit{loss-based attacks} simply measure the model’s negative log-likelihood on the sample, assuming that trained-on text receives lower loss than unseen text. \textit{Reference-based attacks} compare the target model’s loss to that of a second reference model, aiming to factor out distributional biases. A \textit{Zlib-based} technique normalizes loss values by the input’s compressed size, under the rationale that more “compressible” texts may yield different loss profiles for training members versus outsiders. Methods such as \textit{Min-K\%} and \textit{Min-K\%++} focus on particularly low-likelihood tokens, hypothesizing that training members contain fewer extremely unlikely tokens compared to non-members. Although these attacks are designed for a privacy context—detecting memorized examples—our results show that they can fail when non-members are synthetic.

\subsection{Zero-Shot Machine-Generated Text Detection}
\label{sec:background_detectgpt}
Zero-shot detectors for synthetic text (e.g., DetectGPT~\citep{mitchell2023detectgpt}) exploit a model’s probability landscape to expose artifacts of generation. These methods typically insert small, controlled perturbations into candidate text and measure how sharply the likelihood changes. Machine-generated passages often behave like local maxima in the model’s probability space, so analyzing the curvature around these maxima can reveal synthetic origins without labeled examples. Variants such as Fast-DetectGPT~\citep{bao2024fast} streamline this approach by limiting perturbations or using more efficient scoring procedures.

\subsection{Why MIAs Can Function as Machine-generated Text Detectors}
\label{sec:connection}
MIAs and text detectors both hinge on signals derived from the target model’s internal probabilities, especially in their perturbation-based or likelihood-focused forms. Synthetic text, by construction, occupies high-probability regions of a language model’s distribution and can thus elicit membership-like scores when evaluated with MIAs. In other words, an attack designed to flag “memorized” data can easily conflate “machine-generated” with “machine-memorized,” because the underlying scoring mechanism was never intended to separate one model’s synthetic outputs from another model’s genuine training set. As shown in our experiments, this confusion can lead to drastically misleading conclusions regarding model memorization and privacy leakage when non-members are replaced by synthetic text.

\section{Experimental Setup}
\label{sec:experiments}

We evaluate whether MIAs mistakenly treat machine-generated text as training data by comparing two main setups: a \textit{conventional} one that contrasts human-written members and non-members, and a \textit{synthetic} one that replaces the human non-members with generated continuations. Figure~\ref{fig:methodology} offers a visual overview of this process.

\subsection{Data}
\label{sec:data}
We use the MIMIR benchmark~\citep{duan2024membership}, focusing on Wikipedia and ArXiv subsets where membership labels (in/out of training) are verified. These subsets also mitigate high $n$-gram overlaps between members and non-members. In the \textbf{synthetic setup}, we prompt generator models with the first 30 tokens of each human non-member and produce continuations up to 200 tokens, following the method of \citet{mitchell2023detectgpt}. This yields a pool of synthetic non-members from diverse generators, including LLaMA 2, LLaMA 3.1, GPT-3.5, and others.

\subsection{Target Model and Attacks}
Our primary target model is GPT-Neo 2.7B, which is well-documented and trained on public data~\citep{pile}. We apply five different MIAs (loss-based, reference-based, Zlib, Min-K\%, Min-K\%++) as described in Section~\ref{sec:background_mia}, using recommended hyperparameters and code from prior work~\citep{duan2024membership}.

\subsection{Evaluation Protocol}
Each experiment involves three pools of data: (1) \textit{human-written members} from the MIMIR benchmark, (2) \textit{human-written non-members}, and (3) \textit{synthetic non-members} generated as above (Sec.~\ref{sec:data}). For the \textbf{conventional} setup, we evaluate membership attacks on (1) vs.\ (2). For the \textbf{synthetic} setup, we keep the same set of members (1) but replace non-members (2) with synthetic (3). We quantify performance with the area under the ROC curve (AUC-ROC). A dramatic drop in performance (often below random guessing) when moving to the synthetic setup confirms that MIAs systematically misclassify machine-generated text as training data.

\begin{table}[h!]
\centering
\caption{MIA Performance Comparison Across Different Data Sources and Attacks}
\resizebox{\textwidth}{!}{ % Resize the table to fit within the text width
\begin{tabular}{@{}llcccccccccc@{}}
\toprule
\label{tab:results}
\rotatebox{90}{} &  & \multicolumn{5}{c}{Wikipedia} & \multicolumn{5}{c}{ArXiv} \\ 
\cmidrule(lr){3-7} \cmidrule(lr){8-12}
 & Non-members & LOSS & min-k & min-k++ & Ref & zlib & LOSS & min-k & min-k++ & Ref & zlib \\ 
\midrule
\multirow{1}{*}{} & Human-written & 0.657 & 0.650 & 0.637 & 0.606 & 0.623 & 0.790 & 0.760 & 0.655 & 0.718 & 0.784 \\
\cmidrule(lr){1-12}
\multirow{5}{*}{\rotatebox{90}{\textbf{Synthetic}}} 
& GPT-Neo 2.7B & 0.238 & 0.080 & 0.024 & 0.061 & 0.687 & 0.326 & 0.044 & 0.034 & 0.430 & 0.928 \\
& Pythia & 0.309 & 0.163 & 0.116 & 0.409 & 0.799 & 0.457 & 0.127 & 0.084 & 0.761 & 0.940 \\
& Llama 2-7B & 0.431 & 0.340 & 0.269 & 0.559 & 0.958 & 0.694 & 0.474 & 0.489 & 0.908 & 0.996 \\
& Llama 3.1-8B & 0.198 & 0.169 & 0.126 & 0.593 & 0.746 & 0.324 & 0.251 & 0.362 & 0.924 & 0.931 \\
& GPT-3.5 & 0.387 & 0.332 & 0.262 & 0.613 & 0.650 & 0.613 & 0.457 & 0.534 & 0.892 & 0.909 \\
\bottomrule
\end{tabular}
}
\end{table}

\section{Experimental Results}
\label{sec:experimental_results}

Table~\ref{tab:results} presents the performance of various MIAs on two datasets (Wikipedia and ArXiv), comparing a conventional setup with human-written non-members against a synthetic setup in which non-members are generated by multiple models. We report the AUC (area under the ROC curve), which ideally should exceed 0.5 when the attack accurately distinguishes training members from non-members.

\paragraph{Misclassification of Synthetic Text as Training Data.}
Observe that many attacks drop to \emph{well below} random chance (AUC $\ll 0.5$) when confronted with synthetic non-members. For instance, under the Wikipedia subset, \textbf{LOSS}, \textbf{min-k}, and \textbf{min-k++} all plummet to under 0.25 AUC when GPT-Neo 2.7B itself produces the synthetic text. This indicates that the MIA confuses machine-generated continuations with genuine training samples, effectively \emph{reversing} membership predictions: synthetic text is scored as more ``member-like’’ than actual members. As discussed in Section~\ref{sec:connection}, this phenomenon arises because MIAs and text detectors rely on similar likelihood signals; text generated by a language model tends to inhabit high-probability regions in that same model’s distribution, fooling the MIA.

\paragraph{Cross-Model Transfer.}
The issue is particularly striking when the \emph{same} model that is being attacked (GPT-Neo 2.7B) also generates the synthetic text: certain attacks such as \textbf{min-k\%++} yield AUC values near 0.0, suggesting the attack treats these synthetic samples as if they were highly memorized. Moreover, the pattern remains dire even across model boundaries. For instance, synthetic text from GPT-3.5 or LLaMA 3.1 also severely disrupts MIAs on GPT-Neo. This cross-model transfer implies that if non-members are replaced with synthetic text from any large language model—\emph{even one with a different architecture or training corpus}—the MIA can be thoroughly misled.

\paragraph{Implications for Evaluation Protocols.}
These findings have far-reaching consequences. If evaluations rely on synthetic or machine-translated text to approximate “unseen” data, membership analyses may become essentially invalid, as the MIA’s apparent performance may reflect its ability to detect machine-generated text rather than genuine training leakage. Such pitfalls become especially problematic in real-world scenarios where synthetic text proliferates online and can be inadvertently picked up as “non-member” data in future LLM assessments.

\paragraph{Zlib as an Outlier.}
An intriguing exception is the \textbf{zlib} attack, which frequently remains above 0.5 AUC even under synthetic settings (e.g., 0.958 AUC on Wikipedia for LLaMA 2-7B, and 0.996 on ArXiv). This outlier behavior suggests that normalizing by compression size circumvents certain artifacts that plague purely likelihood-based approaches, although the exact reason for this resilience warrants further exploration.

\section{Discussion and Future Work}
\label{sec:discussion_and_future_work}

Our results reveal that many membership inference attacks unintentionally act as \textit{machine-generated text detectors}, thereby undermining their intended purpose of identifying training set membership. Below, we outline the key takeaways, wider implications, and directions for future investigation.

\paragraph{Key Observations and Takeaways.}
\begin{itemize}
    \item \textbf{Synthetic Text Biases MIAs.} When human-written non-members are replaced by synthetic counterparts, most MIA performance metrics plummet below random guessing. The attacks mistakenly interpret machine-generated text as highly “member-like,” calling into question any memorization conclusions drawn under such conditions.
    \item \textbf{Cross-Model Transfer Exacerbates the Problem.} This issue persists even when the generating model differs from the target model. Translated, paraphrased, or otherwise model-produced text could similarly cause MIAs to fail, making it critical to avoid synthetic data in membership evaluations.
    \item \textbf{Zlib Attack Stands Out.} The zlib-based approach remains more robust to synthetic artifacts, suggesting compression-based normalization may mitigate some confounding factors that purely likelihood-based methods cannot.
\end{itemize}

\paragraph{Implications for LLM Evaluation.}
As large language models become more prevalent, synthetic text is increasingly widespread—whether as content on the web or as part of data augmentation pipelines. This poses a grave risk for membership inference research and any related task that relies on comparing “real” vs.\ “unseen” examples. If future evaluations unknowingly incorporate synthetic or model-generated text as a stand-in for non-members (e.g., to sidestep copyright or privacy concerns), the resulting analyses risk conflating machine-generatable text with truly memorized data. Moreover, as LLMs themselves are used for tasks like benchmarking other LLMs, this confusion may propagate into downstream evaluations of creativity, originality, or generalization, all while ignoring the synthetic bias.  

\paragraph{Future Work.}
Building on these insights, several avenues emerge:
\begin{itemize}
    \item \textbf{Redesigning Non-Member Selection.} Curating genuine human-authored non-members—free of temporal or distributional shifts—may necessitate new data-collection frameworks or collaborative agreements to ensure realism and diversity without contamination by synthetic text.
    \item \textbf{Developing Robust MIAs.} Crafting attacks (or modifications to existing ones) that remain reliable in the presence of synthetic text is critical. The zlib attack’s outlier success hints at broader strategies, such as compression-based or hybrid normalization, for distinguishing high-likelihood text from actual memorized samples.
    \item \textbf{Investigating Model Reliance on Synthetic Artifacts.} Understanding \emph{why} machine-generated text so effectively mimics memorized text could lead to new insights into language model probability landscapes, tokenization schemes, and sampling biases.
    \item \textbf{Mitigating Synthetic Overlap.} As synthetic content floods the web and inevitably appears in training corpora, investigating how repeated exposure to model-generated text affects future generations (and subsequent membership evaluations) is an increasingly important concern.
\end{itemize}

In conclusion, our study illuminates a fundamental pitfall in MIA evaluation: synthetic data is not a reliable substitute for genuine non-member examples. Researchers should exercise caution when using machine-generated or translated text as a proxy for out-of-training-distribution data, lest they draw unwarranted conclusions about memorization or data leakage. By recognizing and addressing these overlaps between membership inference and text detection, we can steer future language model evaluations toward greater robustness and interpretability.

%\section*{References}
\bibliographystyle{plainnat}
\bibliography{neurips_2024}
\begin{comment}

\end{comment}

\end{document}